\documentclass[12pt,a4paper]{article}
\usepackage{authblk}
\usepackage[sort&compress,numbers]{natbib}
\usepackage{helvet}
\usepackage{courier}
\usepackage{graphicx}
\usepackage{amsmath,amssymb}
\usepackage{graphicx}
\usepackage[T1]{fontenc}
\usepackage[utf8]{inputenc}
\usepackage{url}
\usepackage{microtype}
\usepackage{units}

\usepackage[plain]{fancyref}
\usepackage{graphicx}
\usepackage{multirow}
\usepackage{fullpage}

\graphicspath{{./graphs/}}

\begin{document}

\title{Localized Coulomb Descriptors for the Gaussian Approximation Potential}

\author[1,2]{James Barker}
\author[1]{Johannes Bulin}
\author[1]{Jan Hamaekers}
\author[1]{Sonja Mathias}
\affil[1]{{\small Fraunhofer-Institut f\"ur Algorithmen und Wissenschaftliches Rechnen SCAI, Schloss Birlinghoven, D-53754 Sankt Augustin}}
\affil[2]{{\small Institut f\"ur Numerische Simulation, Universit\"at Bonn, Wegelerstr. 6, D-53115 Bonn}}

\date{}
\maketitle

\begin{abstract}
  We introduce a novel class of localized atomic environment
  representations, based upon the Coulomb matrix. By combining these
  functions with the Gaussian approximation potential approach, we
  present LC-GAP, a new system for generating atomic potentials
  through machine learning (ML). Tests on the QM7, QM7b and GDB9
  biomolecular datasets demonstrate that potentials created with
  LC-GAP can successfully predict atomization energies for molecules
  larger than those used for training to chemical accuracy, and can
  (in the case of QM7b) also be used to predict a range of other
  atomic properties with accuracy in line with the recent
  literature. As the best-performing representation has only linear
  dimensionality in the number of atoms in a local atomic environment,
  this represents an improvement both in prediction accuracy and
  computational cost when considered against similar Coulomb
  matrix-based methods.
\end{abstract}

\section{Introduction}

An important problem in computational materials science is the
development of efficient and accurate \emph{atomic potentials}. For an
$N$-particle atomic system, an atomic potential (or simply
\emph{potential}) is a function $E^{(N)} : \mathbb{R}^{3N} \times
\mathbb{Z}^N \to \mathbb{R}$ that maps from the set of
three-dimensional Cartesian coordinates locating the atoms in that
system, and their corresponding atomic numbers, to a potential energy.

The most accurate possible potential for any given system can be
obtained from an (analytic) wavefunction solution to Schr\"odinger's
equation \cite{griebel2007}, and all other potentials can be viewed as
approximations to the same. Such solutions must be approximated
numerically in all but the simplest of cases, a process which becomes
computationally expensive for even relatively small systems. A variety
of \emph{first-principle} methods have been developed for this task,
including most notably the density functional theory (DFT)
\cite{tadmor2011}; these methods can be used to obtain potentials
displaying high levels of accuracy, but are nevertheless always
computationally expensive.

By contrast, empirical potentials are parametrized functions, with a
fixed functional form that is usually physically motivated
\cite{plimpton2012}. The parameters of an empirical potential must be
fitted against the properties of a particular material, which can be a
difficult task requiring a level of intuition; once fitted, however,
the evaluation of an empirical potential for an arbitrary atomic
configuration is comparatively cheap. Although it is difficult to
accurately reproduce material properties other than those to which the
potential was fitted, limiting their use in general cases, their
applicability to the problem of exploring the energy hypersurface
immediately surrounding an atomic configuration makes them useful for
simulations.

In recent years, machine learning (ML) approaches have also been
applied to the problem of potential generation. Such approaches can
capture, or ``learn'' information from a wider range of materials than
standard empirical potentials. The resulting ML potentials are almost
as fast to evaluate as empirical potentials, and retain acceptable
accuracy when used to predict the energy of a larger variety of materials.

Notable ML algorithms that have already been used for generating
potentials include kernel ridge regression \cite{hansen2013,
  hansen2015, rupp2015, rupp2012} and multilayer neural networks
\cite{montavon2012}. Another promising approach is the Gaussian
approximation potential (GAP) by Bart\'ok et al. \cite{bartok2010},
which will be explained in detail in \Fref{sec:VMD_methods}. This
potential requires the construction of intermediate representations of
the \emph{atomic neighborhoods} surrounding each atom in the system;
the choice of this representation is crucial for the performance of
the GAP. Bart\'ok originally proposed a modified bispectrum method for
producing descriptors of mono-species crystalline
environments. Although this approach performs well for semiconductors
with respect to accuracy, the nature of the representation as
coefficients of an expansion in spherical harmonics makes it costly to
evaluate. Subsequently, Bart\'ok et al. replaced the bispectrum
representation with an approach called the \emph{SOAP kernel}, which
involves expanding a density function associated with an atomic
environment in spherical harmonics and to directly define the
similarity between any two such density functions~\cite{bartok2013}.

Here, we present an alternative approach to the encoding of atomic
environments, which we refer to as the \emph{localized Coulomb} (LC)
representation. This representation is based on the Coulomb matrix,
originally proposed by Rupp et al. \cite{rupp2012}, and its
derivative, the sorted Coulomb matrix
\cite{montavon2012, rupp2015b}. We combine the LC representation with
the Gaussian approximation potential to obtain a new method for the
creation and evaluation of families of atomic potentials; we term this
method \emph{LC-GAP}.

The remaining article is organized as follows.  In
\Fref{sec:VMD_methods} we briefly summarize the basics for machine
learning for potential energy surfaces and describe our new Coulomb
matrix based descriptors in particular in \Fref{sec:VMD_Coulomb}.  In
\Fref{sec:VMD_results} we give numerical results for several datasets
and additionally discuss the distribution of individual atomic
contributions in \Fref{sec:VMD_AtomicDist}. We conclude with some
remarks in \Fref{sec:VMD_conclusions}.

\section{Potential Energy Prediction through Machine Learning}
\label{sec:VMD_methods}

As stated above, our goal is to produce a function $E^{(N)} :
\mathbb{R}^{3N} \times \mathbb{Z}^N \to \mathbb{R}$ that approximates
the potential energy for an arbitrary unknown atomic system $X^* \in
\mathbb{R}^{3N} \times \mathbb{Z}^N$. Given a collection of atomic
systems of size $N$ and their energies, $\{(X_i, E_i)\}_{i=1}^M$ with
$X_i \in \mathbb{R}^{3N} \times \mathbb{Z}^N$, an arbitrary ML
algorithm could be used to produce, or \emph{learn} the function
$E^{(N)}$. However, such an approach will not scale well to larger
systems; the dimension of $E^{(N)}$ grows linearly in the number of atoms,
and the so-called ``curse of
dimensionality'' implies that the computational cost of learning will
grow exponentially in the same.

Additionally, the use of raw representations $X_i$ as inputs for an ML
process is problematic from a physical standpoint. The potential
energy of any given molecule is necessarily invariant under
translation and orthogonal transformation; it is also invariant under
permutation of the order in which atoms are described
\cite{tadmor2011}. A raw representation obeys none of these
restrictions; therefore, a single unique system can be described by
uncountably many raw representations. This offers a significant
obstacle to the learning and prediction abilities of any ML algorithm.

The Gaussian approximation potential (GAP) framework of Bart\'ok et
al. addresses both these problems \cite{bartok2010}. The first, by
introducing an ansatz that can be used to limit the dimensionality of
a representation, and the second, by using more sophisticated
representations that maximize invariance.  In this section, we briefly
recap the GAP framework, and present some alternative, simpler,
representations of atomic systems.

\subsection{The GAP Framework and Gaussian Process Regression}
\label{sec:VMD_GAP}

The fundamental assumption of the GAP framework is the \emph{atomic
decomposition ansatz}: that the potential energy of an atomic system
can be written as the sum of energies attributed to each of its atoms,
and these atomic energies depend only on a neighborhood of the
corresponding atom. For systems without long range electrostatic
interactions, this assumption can be motivated by
the \emph{nearsightedness of electronic matter} \cite{kohn1996,
prodan2005}.  This allows us to write the potential energy of an
$N$-particle system $X\in\mathbb{R}^{3N}\times \mathbb{Z}^N$ as a sum
of atomic energy contributions:
\begin{equation}
    \label{eq:VMD_GAP}
    E^{(N)}\left(X\right) = \sum_{i=1}^{N}
E_{\mathrm{atomic}}\left(L_i^{(N)}\left(X\right)\right).
\end{equation}
Here, the $L^{(N)}_i(X) :
\mathbb{R}^{3N}\times\mathbb{Z}^N\to\mathbb{R}^d$ are \emph{atomic
  neighborhood representation functions}. Each of these maps a full
atomic system to a $d$-dimensional representation, or
\emph{descriptor}, of the atomic neighborhood of the $i$th
atom. $E_{\mathrm{atomic}} : \mathbb{R}^d \to \mathbb{R}$ is then an
unknown function that assigns an atomic energy contribution to an
atomic neighborhood representation. Given \Fref{eq:VMD_GAP} above, the
problem of learning the function $E^{(N)}$ reduces to the problem of
learning the function $E_{\mathrm{atomic}}$. This function is
necessarily dependent on the form of the functions $L^{(N)}_i$ and the
chosen dimensionality of their output $d$, both of which will be
discussed in the next section.

As originally presented, the GAP framework uses the method of Gaussian
process regression (GPR) to learn the function $E_{\mathrm{atomic}}$,
although other suitable ML algorithms could also be used. The input to
the learning process is a \emph{training set} of $M$ systems, each
containing some number of atoms $ {\textstyle\{N_i\}_{i=1}^M}$, and their known
energies: $ {\textstyle X = \left\{\left(X_i \in \mathbb{R}^{3N_i} \times
  \mathbb{Z}^{N_i}, E_i \in \mathbb{R}\right)\right\}_{i=1}^M}$. The ability
to learn from a training set containing systems with differing numbers
of atoms is a direct result of learning the function
$E_{\mathrm{atomic}}$ rather than $E_{\mathrm{total}}$, the energy of the
whole molecule. Additionally,
a level of Gaussian observation noise $\varepsilon_i
\sim\mathcal{N}\left(0, \sigma_i^2\right)$ is associated with each
training set energy $E_i$, such that
\begin{equation*}
  E_i = \left(\sum_{j=1}^N
E_{\mathrm{atomic}}\left(L^{(N)}_j\left(X\right)\right)\right) +
\varepsilon_i.
\end{equation*}
Gaussian process regression predicts the function
$E_{\mathrm{atomic}}$ as a linear combination of positive-definite
kernel functions $\kappa : \mathbb{R}^d \times \mathbb{R}^d
\to \mathbb{R}$ centered on the atomic neighborhood representations of
each atom in every system in the training set. That is, for some
$N$-particle system $X^*$ whose energy we wish to predict,
\begin{equation*}
  E_{\mathrm{atomic}}\left(L^{(N)}_k(X^*)\right) = \sum_{i=1}^M\alpha_{i}\sum_{j=1}^{N_i}
\kappa\left(L^{(N_i)}_j\left(X_i\right),
L^{(N)}_k\left(X^*\right)\right),
\end{equation*}
where the $\alpha_{i}$ are chosen during the learning process. Again,
note that the use of $E_{\mathrm{atomic}}$ rather than
$E_{\mathrm{total}}$ allows the prediction of potential energies for
systems with arbitrary numbers of atoms, regardless of the contents of
the training set.

\subsection{Localized Coulomb Matrix Descriptors}
\label{sec:VMD_Coulomb}

The choice of representation functions $ {\textstyle L_i^{(N)} : \mathbb{R}^{3N}
\times \mathbb{Z}^N \to \mathbb{R}^d}$ is critical to the behavior of
the GAP framework, both in terms of computational performance and in
accuracy. The primary contribution of this paper is the introduction
of three new such functions, which are described below.

The genesis of our work is the Coulomb matrix descriptor, due to Rupp
et al. \cite{rupp2012}, which allows the representation of complete
molecules as square matrices. Consider an $N$-atom system
$X\in\mathbb{R}^{3N}\times \mathbb{Z}^N$, consisting of Cartesian
coordinates $\textbf{R}_1,\dots,\textbf{R}_N$ describing atom
locations, and the atomic numbers $Z_1,\dots, Z_N$ associated with
those atoms. The Coulomb matrix is then an $N\times{}N$ matrix $M$,
whose entries are given as
\begin{equation}
  \label{eq:VMD_Coulomb}
  M_{ij} = \begin{cases} 0.5 Z_i^{2.4} & \hspace{5pt} i=j,\\[3pt]
    \frac{Z_iZ_j}{\|\textbf{R}_i-\textbf{R}_j\|_2} &\hspace{5pt} i \neq
j.
  \end{cases}
\end{equation}
By construction, the Coulomb matrix is invariant under translations
and orthogonal transformations of the set of Cartesian coordinates
${\textstyle \{\textbf{R}_1, \ldots, \textbf{R}_N\}}$. It is, however,
not invariant under permutation of the indexing order of the
coordinates and their associated atomic numbers.  To improve the
handling of permuted systems, the rows and columns of the Coulomb
matrix can be sorted according to their respective norms (which are
identical, as the matrix is symmetric), resulting in the \emph{sorted
  Coulomb matrix} \cite{montavon2012}.

The sorted Coulomb matrix is a \emph{global descriptor} of an atomic
system. As such, it can only represent molecules, and is inapplicable
to infinitely-periodic crystal systems. However, it is possible to
modify the Coulomb matrix into a \emph{local descriptor}, encoding
information only about the immediate neighborhood of an atom. Such a
descriptor is then a candidate for unqualified use as an atomic
representation function $L_i^{(N)}$ in the GAP framework, and can be
applied to infinitely-periodic crystal systems as well as finite
molecules.

Before we begin, it is useful to define a system of \emph{local
indices} to specify more clearly the structure of an atomic
neighborhood around the $i$th atom in some system $X$. First, let
$p_1$ indicate the $i$th atom itself. Then let $K$ be the number of
atoms that are located within some cutoff radius $r_{\mathrm{cut}} >
0$ around $\mathbf{R}_i$, and let $p_2, \ldots{}, p_{K+1}$ specify
these atoms. Finally, let $m$ be the maximum neighborhood occupancy,
which must be chosen in such a way that it is impossible to have
greater than $m-1$ atoms surrounding any atom in either the training
set, or an atom in a system which will be predicted. Then let all
indices $p_{K+2}, \ldots{}, p_m$ specify \emph{dummy atoms}, which
have arbitrary location and atomic number zero. The use of such dummy
atoms is an established procedure \cite{montavon2012}, and serves to
ensure that all descriptors produced in either the training or
prediction process will be the same size and therefore comparable.

With the aid of local indices, we now define the \emph{localized
Coulomb matrix} for the $i$th atomic neighborhood. First, similarly
to \Fref{eq:VMD_Coulomb} above, define the $m \times{} m$ matrix $M^{(i)}$,
with entries given by
\begin{equation}
\label{eq:VMD_localized}
    M^{(i)}_{jk} =  \begin{cases}
        0.5 Z_{p_j}^{2.4} & \hspace{5pt} j=k,\\[3pt]
        \frac{Z_{p_{j}}
Z_{p_{k}}}{\|\textbf{R}_{p_{j}}-\textbf{R}_{p_{k}}\|_2^\alpha}
&\hspace{5pt} \text{otherwise}.
    \end{cases}
\end{equation}
(Note that for notational simplicity, we do not explicitly indicate
the dependency of the $\left(p_1, \ldots, p_m\right)$ local indices on
the choice of $i$.) Then let $P$ be the permutation matrix such that
$P_{1,1} = 1$ and
\begin{equation}
    \left\|\left(P M^{(i)} P^T\right)_{j,\ast}\right\|_2 \geq \left\|\left(P M^{(i)} P^T\right)_{j+1,\ast}\right\|_2
\end{equation}
for $j=2,\ldots,m-1$. That is, $P$ reorders the rows and columns of
$M^{(i)}$ such that those corresponding to the central atom $p_1$
become the upper- and leftmost respectively, while permuting the
remainder in descending order of norm. The permuted matrix
$PM^{(i)}P^T$ is called $C^{(i)}_{\textrm{loc}}$ and is termed the \emph{localized
Coulomb matrix} for the $i$th neighborhood. Finally, the upper
triangular entries of this matrix are packed in row-wise order into a
vector $\mathbf{c}$ of size $d=\frac{m(m+1)}{2}$, and $L_i^{(N)}(X)
:= \mathbf{c}$ is the \emph{localized Coulomb descriptor
function}. (Because the matrix is symmetric, the lower-triangular
entries contain no further information.)

One issue with this choice of representation function is that small
changes to the state of the system that result in atoms entering or
leaving the cutoff radius can cause large changes to the resulting
descriptor. To reduce the significance of this effect, we modify the
denominator to include distances from each atom to the central atom,
creating the \emph{decaying Coulomb matrix} $C^{(i)}_{\mathrm{dec}} =
P\hat{M}^{(i)}P^T$, with:
\begin{equation}
  \label{eq:VMD_decaying}
  \hat{M}^{(i)}_{jk} =  \begin{cases}
    0.5 Z_{p_j}^{2.4} & \hspace{5pt} j=k=1,\\
    \frac{Z_{p_j} Z_{p_k}}{\left(
        \left\|\textbf{R}_{p_1} - \textbf{R}_{p_j}\right\|_2
        + \left\|\textbf{R}_{p_1} - \textbf{R}_{p_k}\right\|_2
        + \left\|\textbf{R}_{p_j}-\textbf{R}_{p_k}\right\|_2\right)^\alpha}
&\mbox{ otherwise}.
  \end{cases}
\end{equation}
The permutation matrix is chosen in the same fashion as for the
localized Coulomb matrix. Due to the introduction of the additional
distance terms, the movement of atoms across the cutoff boundary cause
significantly smaller changes, which may be lower than machine
precision for some values of $r_{\textrm{cut}}$ and $\alpha$. The
decaying Coulomb matrix can be packed into a vector of size
$d=\frac{m(m+1)}{2}$ in the same manner as before.

Finally, we introduce a lower-dimensional descriptor, called the
\emph{reduced Coulomb matrix}, based in turn upon the decaying Coulomb
matrix. Rather than use a permutation matrix, we require that the
atoms $p_2,\dots, p_{k+1}$ are indexed in such a way that
$\|\textbf{R}_{p_j} - \textbf{R}_{p_1}\|_2\leq \|\textbf{R}_{p_{j+1}} -
\textbf{R}_{p_1}\|_2$ for $j=1,\dots,m-1$, similar to the approach in
\cite{rupp2015b}.  Then the descriptor is constructed by taking the
first row and the diagonal of the matrix $\hat{M}^{(i)}$
in \Fref{eq:VMD_decaying}:
\begin{equation}\label{eq:VMD_reduced}
  C^{(i)}_{\mathrm{red}} = \left[\hat{M}^{(i)}_{11}, \hat{M}^{(i)}_{12}, \ldots, \hat{M}^{(i)}_{1m}, \hat{M}^{(i)}_{22},
\hat{M}^{(i)}_{33}, \ldots, \hat{M}^{(i)}_{mm} \right].
\end{equation}
This descriptor has dimensionality $2m-1$, linear rather than
quadratic in the maximum neighborhood occupancy, while still encoding
all information about the pairwise interaction of the central particle
with all its neighboring atoms.

\section{Results}
\label{sec:VMD_results}

In order to evaluate the new atomic neighborhood functions presented
in \Fref{sec:VMD_Coulomb} above, we performed a series of numerical
experiments over the QM7, QM7b, and GDB9 datasets. The QM7
dataset~\cite{blum2009, rupp2012} contains approximately 7100
biomolecules selected from the GDB13 database, each with up to seven
heavy atoms\footnote{All non-hydrogen atoms are considered heavy.},
and their atomization energies\footnote{The atomization energy is the
potential energy of a molecule that has been adjusted by the combined
potential energy of its isolated atoms \cite{filszar1994}.},
calculated at the PBE0 level of theory. The QM7b
dataset~\cite{blum2009, montavon2013} is a slightly-expanded version of
QM7, containing 7211 biomolecules with up to seven heavy atoms,
including some with chlorine. As well as atomization energy, QM7b
provides a total of thirteen extra properties per molecule
(e.g. polarizability, HOMO and LUMO eigenvalues, excitation energies)
calculated at different levels of theory (ZINDO, SCS, PBE0,
GW). Finally, the GDB9 dataset~\cite{ramakrishnan2014} contains
approximately $134,000$ biomolecules (also selected from GDB13)
containing up to nine heavy atoms, along with their atomization
energies calculated at both the PM7 and B3LYP levels of theory. For
computational convenience, we selected a subset of the full GDB9
database (calculated at PM7), comprised of all molecules with up to
eight heavy atoms, and approximately $18,000$ molecules with nine
heavy atoms, chosen by stratification. We refer to this subset as the
GDB9\_18K dataset.

For each of the three datasets described above, we isolated subsets
containing up to $n$ heavy atoms for all $n \leq m$, where $m$ is the
maximum number of heavy atoms in any molecule in the dataset. We refer
to these subsets as QM7\_1, QM7\_2, QM7\_3, etc. A decomposition of
each dataset by number of heavy atoms per molecule is given
in \Fref{tab:VMD_subset_size}.

\begin{table}[h]
\caption{Decomposition of testing datasets by number of heavy atoms per molecule.}
\label{tab:VMD_subset_size}
{\small
\begin{center}
\begin{tabular}{lrrrrrrrrrr}
\hline
\multicolumn{1}{c}{\multirow{2}{*}{Dataset}}&\multicolumn{9}{c}{Number of heavy atoms}&
\multicolumn{1}{c}{\multirow{2}{*}{Total molecules}}\\
\cline{2-10}
&1&2&3&4&5&6&7&8&9&\\
\hline
QM7&\ \ 1&\ \ 3&\ \ 12&\ \ 43&\ \ 158&\ \ 950&\ \ 5998&--&--&7165\\
QM7b&1&3&12&43&158&953&6041&--&--&7211\\
GDB9-18K&3&5&9&31&130&618&3197&18298&18599&40890\\
\hline
\end{tabular}
\end{center}}
\end{table}

Each experiment consisted of learning a potential function from one
such subset, called the \emph{training set}, and then using the
resulting potential to predict the atomization energies of another
(non-overlapping) subset, called the \emph{test set}. All testing was
performed using an implementation of the Gaussian approximation
potential, using one of the three localized Coulomb representation
functions described in \Fref{sec:VMD_Coulomb} above; we term this
system \emph{LC-GAP}. The underlying ML algorithm used was Gaussian
process regression, as described in \cite{rasmussen2006}.

For these experiments, we used the Laplacian kernel
\begin{displaymath}
    \kappa(x, y) = \sigma^2 \exp\left(\frac{-\left\|x -
y\right\|_1}{l^2} \right),
\end{displaymath}
which has shown good performance when applied to similar problems
\cite{hansen2015,hansen2013}. The terms $\sigma$ and $l$ are kernel
\emph{hyperparameters}, the choice of which can have serious
implications for the accuracy of the system.

To assess the quality of prediction for each test set, we used the
mean absolute error (MAE), defined as
\begin{equation*}
 \textrm{MAE}\left(X^*\right) = \frac{1}{M}\sum_{m=1}^M
\left|E_m^{\textrm{exact}} - E_m^{\textrm{pred}}\right|,
\end{equation*}
where $M$ is the number of entries in the test set, the terms
$E_m^{\textrm{exact}}$ are the atomization energies obtained from the
dataset, and $E_m^{\textrm{pred}}$ the atomization energies obtained
by prediction. This error metric is well-established in the statistics
community for assessing the prediction accuracy of regression models
\cite{willmott2005}, and has been used along with the QM7 dataset to
benchmark other ML-based potentials.

\subsection{Comparison of Descriptor Functions on QM7}

\begin{figure}[h]
\centering
\includegraphics[width=\textwidth]{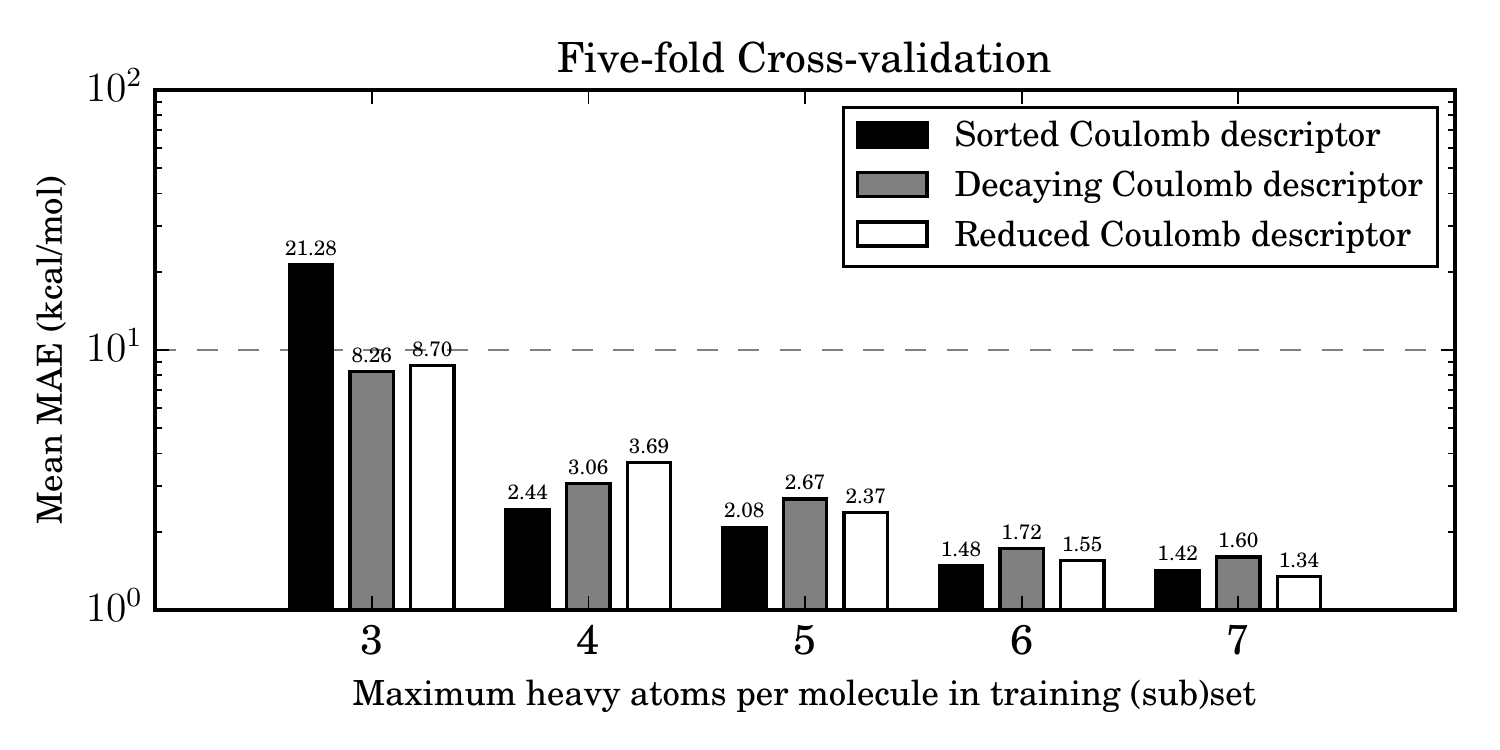}
\caption{Mean MAEs of the predicted atomization energies on various
  QM7 subsets obtained during five-fold cross-validation, using LC-GAP
  potentials equipped with the different localizers described in
  \Fref{sec:VMD_Coulomb}. The precise values of $\alpha$ and
  $r_\textrm{cut}$ used for each result can be found in
  \Fref{tab:cross_validation_parameters_qm7}.}
\label{fig:cross_validation_qm7}
\end{figure}

We began by investigating the performance of the three localized
Coulomb descriptors (sorted, decaying, and reduced) described in
\Fref{sec:VMD_Coulomb} above. To establish the ability of these
descriptors to predict atomization energies of molecules with similar
properties to those used for training, we performed five-fold
cross-validation on each of the QM7 subsets containing at least three
heavy atoms. Kernel hyperparameters ($l$, $\sigma$) were chosen by
minimization of the negative log-likelihood on the training set. For
each descriptor, a grid of localization parameters $\alpha$ and
neighborhood cutoffs $r_\textrm{cut}$ were trialed.\footnote{Here, we
  tested values $(\alpha, r_\textrm{cut}) \in \{3, 4, \ldots
  ,7\}\times \{3.0, 3.5, 4.0, 4.5, \ldots, 7.0\}$.}  A graph
demonstrating the best results obtained on each subset can be found in
\Fref{fig:cross_validation_qm7}; a table describing the values of
$\alpha$ and $r_\textrm{cut}$ for each can be found in
\Fref{tab:cross_validation_parameters_qm7}.

\begin{table}[h]
\caption{Best choices of descriptor parameters ($\alpha$, $r_\textrm{cut}$) for cross-validation over QM7, by descriptor type and data subset. These choices are used to generate the results seen in \Fref{fig:cross_validation_qm7}.}
\label{tab:cross_validation_parameters_qm7}
\begin{center}
\begin{tabular}{llllll}
\hline
Descriptor&QM7\_3&QM7\_4&QM7\_5&QM7\_6&QM7\_7\\
\hline
Localized & (4.0, 5.0) & (3.0, 3.5) & (5.0, 3.5) & (3.0, 3.5) & (3.0, 3.5)\\
Decaying & (3.0, 4.0) & (3.0, 3.5) & (3.0, 3.5) & (4.0, 3.5) & (3.0, 4.5)\\
Reduced & (3.0, 6.0) & (5.0, 6.0) & (3.0, 5.5) & (4.0, 5.5) & (5.0, 6.5)\\
\end{tabular}
\end{center}
\end{table}

\begin{figure}[h]
\centering
\includegraphics[width=\textwidth]{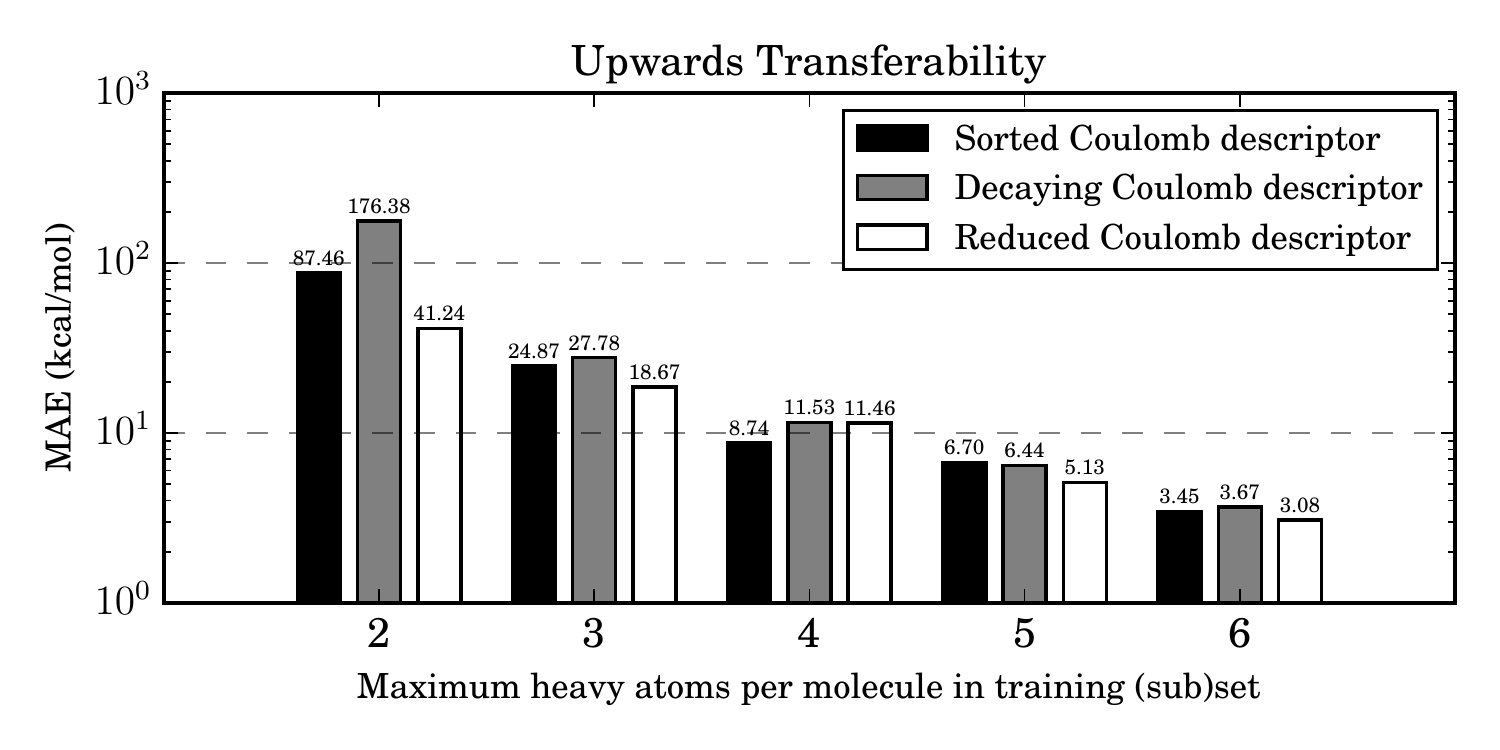}
\caption{MAEs of the predicted atomization energies on the remainder
  of the full QM7, using LC-GAP potentials trained on QM7 subsets,
  equipped with the different localizers described in
  \Fref{sec:VMD_Coulomb}. The precise values of $\alpha$ and
  $r_\textrm{cut}$ used for each result can be found in
  \Fref{tab:upwards_transferability_parameters_qm7}.}
\label{fig:upwards_transferability_qm7}
\end{figure}

\begin{table}[h]
\caption{Best choices of descriptor parameters ($\alpha$,
  $r_\textrm{cut}$) for upwards transferability over QM7, by
  descriptor type and data subset. These choices are used to generate
  the results seen in \Fref{fig:upwards_transferability_qm7}.}
\label{tab:upwards_transferability_parameters_qm7}
\begin{center}
\begin{tabular}{llllll}
\hline
Descriptor&QM7\_2&QM7\_3&QM7\_4&QM7\_5&QM7\_6\\
\hline
Localized& (4.0, 3.5) & (4.0, 3.5) & (4.0, 3.5) & (5.0, 3.5) & (4.0, 4.0) \\
Decaying&(4.0, 3.5) & (4.0, 3.5) & (4.0, 3.5) & (4.0, 4.0) & (5.0, 3.5)\\
Reduced& (4.0, 5.5) & (4.0, 5.5) & (4.0, 5.0) & (5.0, 5.0) & (5.0, 6.0)\\
\end{tabular}
\end{center}
\end{table}

For all but the smallest subset, the three tested descriptor types
perform similarly, all producing results on the same order of
magnitude and (for QM7\_5 and higher) within one kcal/mol of MAE of
each other. Although the localized Coulomb representation performs the
best for the QM7\_4, QM7\_5 and QM7\_6 subsets, it is interesting to
note that on the full QM7 dataset, the reduced Coulomb representation
slightly outperforms both of the others. This is notable, given the
lower dimensionality of that descriptor.

The cross-validation performance of each descriptor allows a direct
comparison with previously-published results in the literature. Hansen
et al. compared several different ML potentials by using five-fold
cross-validation on the complete QM7 dataset \cite{hansen2013}. Using
non-localized kernel ridge regression coupled with randomly-sorted
(global) Coulomb matrices, they report an MAE of 3.07 kcal/mol. Using
multilayer neural networks, they obtained an MAE of 3.51 kcal/mol.
Recently, Hansen et al. also reported an MAE of 1.5 kcal/mol, obtained
through kernel ridge regression combined with the ``Bag of Bonds''
approach \cite{hansen2015}. We note that both the sorted and reduced
Coulomb descriptors outperform this result, with MAEs of 1.42 and 1.34
kcal/mol respectively. We conclude that all three of the tested
descriptors offer competitive accuracy when compared to the
literature.

\begin{table}[h]
\caption{MAE for different algorithms when training and evaluating on the whole QM7 dataset.}
\label{tab:VMD_cross_validation}
\begin{center}
\begin{tabular}{ll}
\hline\noalign{\smallskip}
Algorithm & MAE [kcal/mol] \\
\noalign{\smallskip}\hline\noalign{\smallskip}
LC-GAP (localized Coulomb matrix) & 1.42\\
LC-GAP (decaying Coulomb matrix) & 1.60 \\
LC-GAP (reduced Coulomb matrix) & 1.34 \\
Kernel ridge regression \cite{hansen2013} & 3.07 \\
Multilayer neural network \cite{montavon2012, hansen2013} & 3.51 \\
Bag of Bonds \cite{hansen2015} & 1.5\\
\noalign{\smallskip}\hline\noalign{\smallskip}
\end{tabular}
\end{center}
\end{table}

It is also important to consider the application of an ML-based
potential to datasets with potentially different properties than those
used for training. To evaluate the behavior of the three LC-GAP
localizer functions in such a scenario, we performed ``upwards
transferability'' testing over the QM7 dataset. In these tests, an
LC-GAP potential is trained using one of the QM7 subsets; it is then
used to predict all entries in the remainder of the complete dataset,
which have (by construction) at least one heavy atom more than any
molecule in the training set. The results of these tests are given
in \Fref{fig:upwards_transferability_qm7}.

The achieved MAEs are not as low as in the case of cross-validation,
which is to be expected. However, potentials display consistent
improvement as their training sets increase in complexity; potentials
trained on the QM7\_5 and QM7\_6 datasets achieve MAEs below 10
kcal/mol in all cases. Unlike the cross-validation testing, the
reduced Coulomb descriptor performs the best in all cases except for
QM7\_4. This suggests that despite its lower dimensionality, the
reduced Coulomb descriptor embodies as much ``useful information'' as
the higher-order descriptors, at a significantly lower cost with
respect to both storage and computation.

\subsection{Larger Datasets and Prediction of Multiple Properties}

\begin{figure}[h]
\centering
\includegraphics[width=\textwidth]{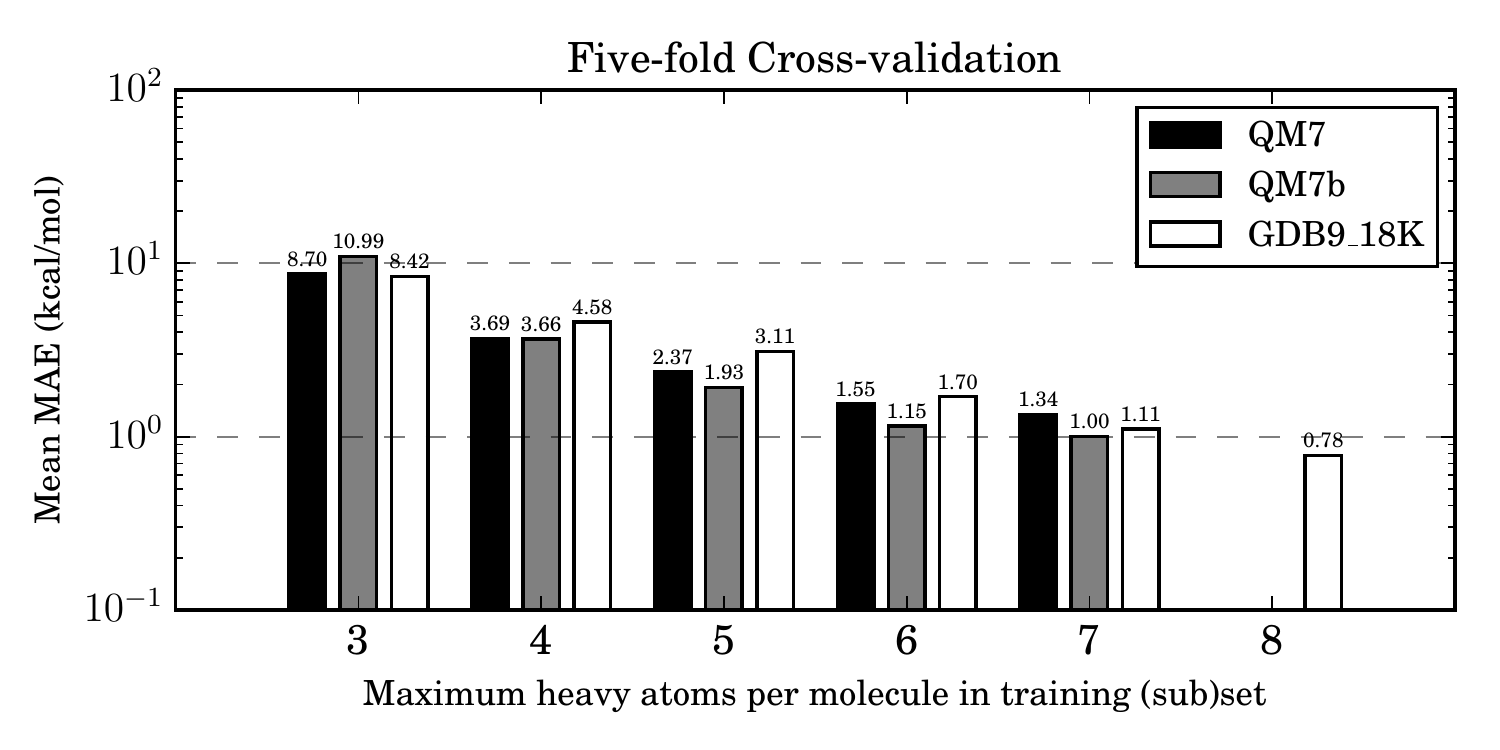}
\caption{Mean MAEs of the predicted atomization energies on subsets of
  QM7, QM7b, and GDB9\_18K, obtained during five-fold
  cross-validation, using LC-GAP potentials equipped with the reduced
  Coulomb descriptor~(\ref{eq:VMD_reduced}). The precise values of
  $\alpha$ and $r_\textrm{cut}$ used for each result can be found in
  \Fref{tab:cross_validation_all_datasets}.}
\label{fig:cross_validation_all_datasets}
\end{figure}

\begin{table}[h]
\caption{Best choices of reduced Coulomb matrix descriptor parameters
  ($\alpha$, $r_\textrm{cut}$) for upwards transferability, by data
  set and number of heavy atoms in training set. These choices are
  used to generate the results seen in
  \Fref{fig:cross_validation_all_datasets}.}
\label{tab:cross_validation_all_datasets}
\begin{center}
\begin{tabular}{lcccccc}
\hline
Dataset&3&4&5&6&7&8\\
\hline
QM7& (3.0, 6.0) & (5.0, 6.0) & (3.0, 5.5) & (4.0, 5.5) & (5.0, 6.5) & -\\
QM7b&(5.0, 3.5) & (5.0, 5.5) & (5.0, 5.5) & (5.0, 5.0) & (5.0, 5.5) & -\\
GDB9&(5.0, 4.5) & (5.0, 4.0) & (5.0, 5.0) & (6.0, 6.0) & (5.0, 4.0) & (5.0, 4.0)\\
\end{tabular}
\end{center}
\end{table}

\begin{figure}[h]
\centering
\includegraphics[width=\textwidth]{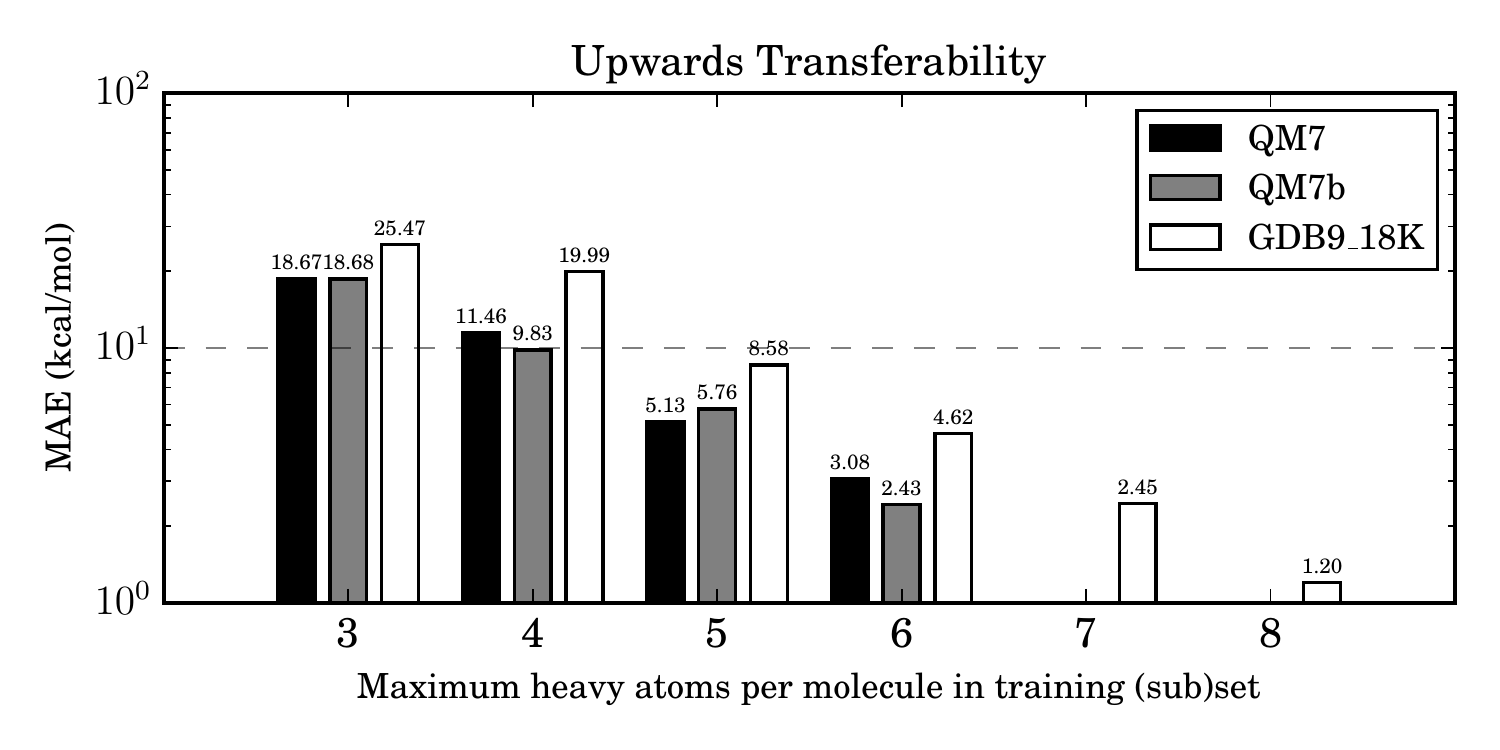}
\caption{MAEs of the predicted atomization energies on the remainder
  of the full dataset, using LC-GAP potentials trained on subsets,
  equipped with the reduced Coulomb
  descriptor~(\ref{eq:VMD_reduced}). The precise values of $\alpha$
  and $r_\textrm{cut}$ used for each result can be found in
  \Fref{tab:upwards_transferability_all_datasets}.}
\label{fig:upwards_transferability_all_datasets}
\end{figure}

To further investigate the performance of LC-GAP, we repeated the
cross-validation and upwards-transferability tests on the atomization
energy figures provided by the remaining two datasets, QM7b and
GDB9\_18K, and their respective subsets. Due to the competitive
performance of the reduced Coulomb descriptor, we did not perform
further testing with the localized and decaying Coulomb descriptors;
similarly, due to computational limitations, we did not perform
cross-validation testing on the entire GDB9\_18K dataset. As before,
kernel hyperparameters were chosen by minimization of the negative
log-likelihood of the training set, and a grid of descriptor
parameters ($\alpha$, $\sigma$) were used. Graphs displaying the
results for each set of experiments can be found
in \Fref{fig:cross_validation_all_datasets}
and \Fref{fig:upwards_transferability_all_datasets} respectively, with
the previously-obtained results for QM7 provided for comparison. The
kernel hyperparameters and descriptor parameters used for each are
given in \Fref{tab:cross_validation_all_datasets}
and \Fref{tab:upwards_transferability_all_datasets}.

\begin{table}[h]
\caption{Best choices of reduced Coulomb matrix descriptor parameters ($\alpha$, $r_\textrm{cut}$) for upwards transferability, by data set and number of heavy atoms in training set. These choices are used to generate the results seen in \Fref{fig:upwards_transferability_all_datasets}.}
\label{tab:upwards_transferability_all_datasets}
\begin{center}
\begin{tabular}{lccccc}
\hline
Dataset&3&4&5&6&7\\
\hline
QM7& (4.0, 5.5) & (4.0, 5.0) & (5.0, 5.0) & (5.0, 6.0) & - \\
QM7b&(4.0, 5.5) & (4.0, 5.0) & (6.0, 5.0) & (5.0, 5.0) & - \\
GDB9&(5.0, 6.0) & (7.0, 4.0) & (5.0, 5.5) & (5.0, 4.0) & (6.0, 4.0)\\
\end{tabular}
\end{center}
\end{table}

In both cross-validation and upwards-transferability tests, the QM7b
results perform similarly, although generally slightly better than the
QM7 results. This is unsurprising, given that the datasets are
similar, and QM7b contains only slightly more molecules than
QM7. However, we note that during cross-validation over the entirety
of the dataset, LC-GAP is able to predict molecules in QM7b to a mean
MAE of approximately 1.00 kcal/mol, i.e., at chemical accuracy.

In contrast, LC-GAP underperforms on GDB9 when considered against both
QM7 and QM7b at the same number of heavy atoms. In the
upwards-transferability case, this can be explained by the fact that
the vast majority of the molecules in GDB9 have eight or nine heavy
atoms, increasing the difficulty of prediction by a potential trained
on molecules containing fewer heavy atoms. For cross-validation, this
is potentially a result of the slightly lower number of molecules in
the smaller subsets (cf. \Fref{tab:VMD_subset_size}). The results for
GDB9 are nevertheless encouraging: a potential trained on molecules
containing eight or fewer heavy atoms can predict the remainder of the
dataset (i.e., all molecules with nine heavy atoms) with an MAE of
only 1.20 kcal/mol, and cross-validation over the entire dataset
produces a mean MAE of 0.78 kcal/mol -- well above chemical accuracy.

\begin{figure}[h]
\centering
\includegraphics[width=\textwidth]{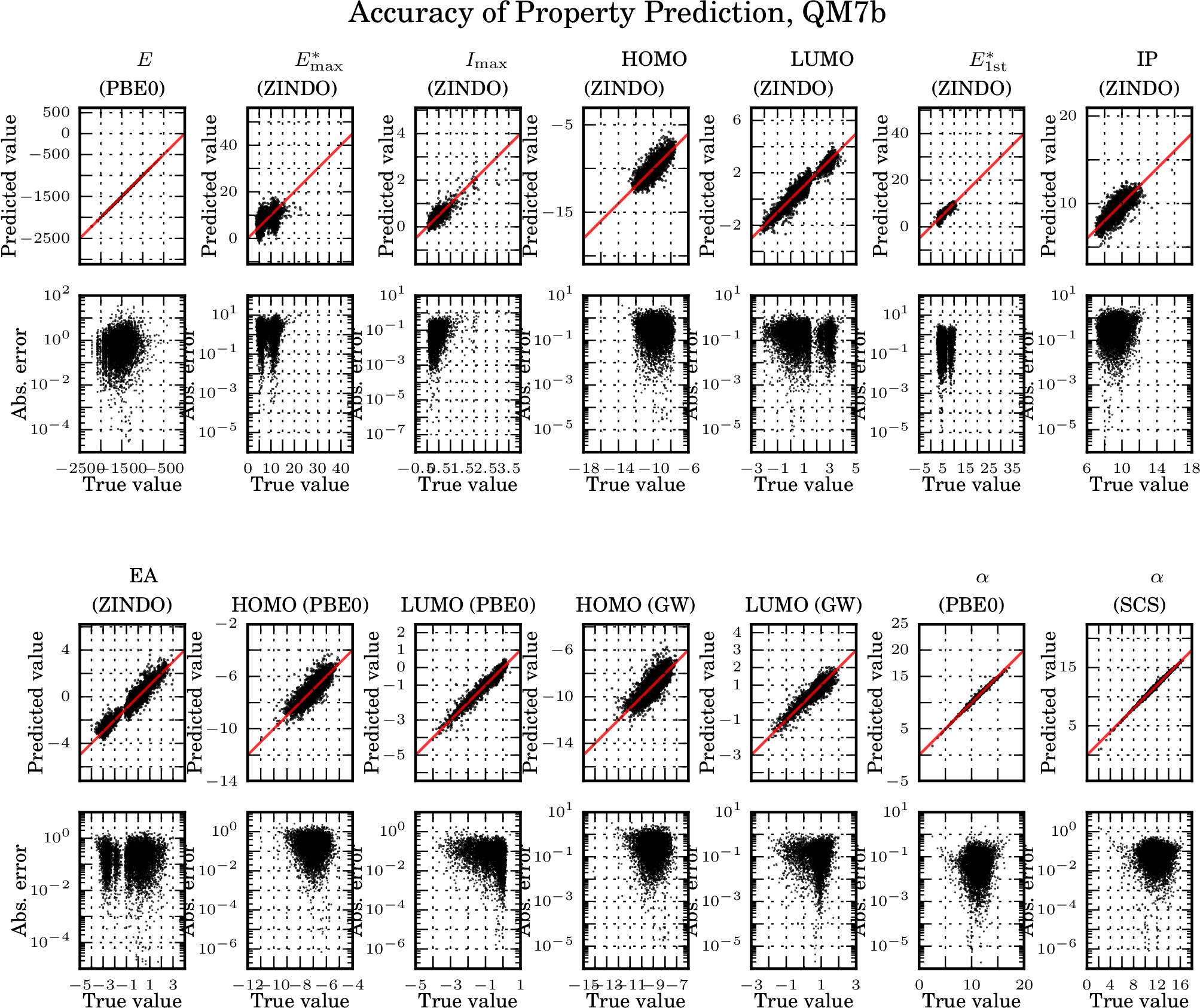}
\caption{Scatter plots displaying the distribution of all results
  obtained during cross-validation over the QM7b dataset for each of
  the fourteen included properties, corresponding to the results in
  \Fref{tab:many_props_cross_validation_results}. In each main row,
  plots in the upper subrow display the obtained prediction values
  plotted against their actual values; plots in the lower subrow
  display the absolute error of the predictions, plotted against the
  actual value.}
\label{fig:many_props_scatters}
\end{figure}

As well as atomization energies, QM7b contains a number of different
molecular properties for each molecule in the dataset, as described
above. In \cite{de2016comparing}, De et al. report prediction results for these
properties obtained with the SOAP kernel. To investigate the
applicability of the LC-GAP system to this problem, we repeatedly
performed cross-validation over the entire QM7b dataset, considering
each property in turn. Importantly, we did not repeat the process of
hyperparameter and descriptor parameter selection for each property;
rather, we used those indicated in the atomization-energy case as
above. A summary of the obtained results and a comparison with those
of De et al. can be found
in \Fref{tab:many_props_cross_validation_results}; additionally,
scatter plots indicating the distribution of both the predicted
results and their absolute errors can be found
in \Fref{fig:many_props_scatters}.

\begin{table}[h]
\caption{Results obtained during cross-validation on the QM7b dataset
  for each of the fourteen molecular properties contained therein,
  with comparison figures obtained from \cite{de2016comparing}. For
  all results, the reduced Coulomb localizer was used; kernel
  hyperparameters and descriptor parameters were equivalent to those
  used to produce the entries in graph for the full QM7b dataset.}
\label{tab:many_props_cross_validation_results}
\begin{center}
\begin{tabular}{
                l @{\hskip 5ex} l @{\hskip 5ex} l @{\hskip 3ex} l @{\hskip 5ex}}
                \multicolumn{1}{l}{\multirow{2}{*}{Property}}
                &\multicolumn{1}{l}{\multirow{2}{*}{Units}}
                &\multicolumn{1}{l}{LC-GAP}
                &\multicolumn{1}{l}{Reference Data \cite{de2016comparing}}\\
                &&\multicolumn{1}{l}{Mean MAE}&\multicolumn{1}{l}{MAE}\\\hline
$E$ (PBE0)                         &kcal/mol     &$1.002 \pm{} 0.022$&$0.92$\\
$E^*_{\textrm{max}}$ (ZINDO)       &eV           &$1.717 \pm{} 0.025$&$1.56$\\
$I_{\textrm{max}}$ (ZINDO)         &arbitrary    &$0.087 \pm{} 0.003$&$0.08$\\
HOMO (ZINDO)                       &eV           &$0.410 \pm{} 0.009$&$0.13$\\
LUMO (ZINDO)                       &eV           &$0.228 \pm{} 0.003$&$0.10$\\
$E^*_{\textrm{1st}}$ (ZINDO)       &eV           &$0.493 \pm{} 0.007$&$0.18$\\
IP (ZINDO)                         &eV           &$0.439 \pm{} 0.010$&$0.19$\\
EA (ZINDO)                         &eV           &$0.267 \pm{} 0.004$&$0.13$\\
HOMO (PBE0)                        &eV           &$0.284 \pm{} 0.006$&$0.11$\\
LUMO (PBE0)                        &eV           &$0.091 \pm{} 0.001$&$0.08$\\
HOMO (GW)                          &eV           &$0.355 \pm{} 0.008$&$0.12$\\
LUMO (GW)                          &eV           &$0.146 \pm{} 0.003$&$0.12$\\
$\alpha$ (PBE0)                    &\AA$^3$       &$0.072 \pm{} 0.002$&$0.05$\\
$\alpha$ (SCS)                     &\AA$^3$       &$0.086 \pm{} 0.002$&$0.02$
\end{tabular}
\end{center}
\end{table}

Without exception, the mean MAE results achieved by LC-GAP are on the
same order of magnitude as those reported in the reference data; many
agree to at least one significant figure, although in no cases are the
results better than those reported; considering the slight difference
between the results for atomization energy, this is unsurprising. The
precise distribution of the results is harder to interpret; almost all
of the lower-subrow scatter plots display similar patterns of errors
to one another, regardless of the difference between the reference
data accuracy and that achieved by LC-GAP. Nevertheless, these results
indicate that the LC-GAP potential is, in general, quite capable of
predicting molecular properties using hyperparameters and descriptor
parameters that were not optimized for that specific property --
potentially a significant benefit, since parameter optimization is
significantly more computationally intensive than simple training and
application of a potential.

\subsection{Distribution of Individual Atomic Contributions}
\label{sec:VMD_AtomicDist}
Through the atomic decomposition ansatz, the GAP approach predicts the
total value of any given property for a molecule as the sum of many
atomic contributions, one for each atom in that molecule. Although
previous work has focused only on the total value of a property, the
atomic contributions are readily available, and are worthy of
interest.

\begin{figure}[h]
\centering
\includegraphics[width=\textwidth]{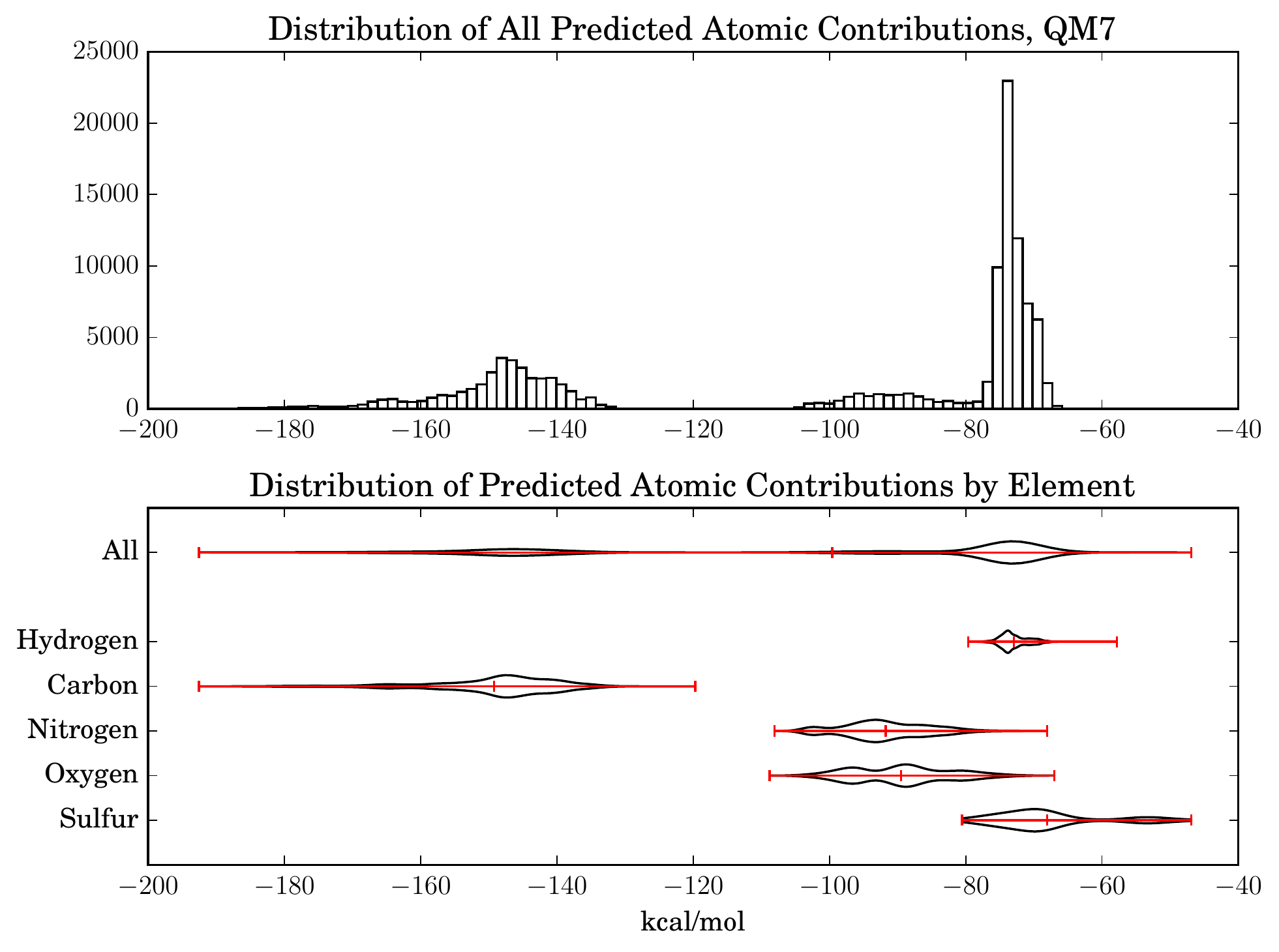}
\label{fig:atomic_contribution_distribution}
\caption{Plots showing the distribution of all calculated individual
  atomic contributions towards the atomization energies of molecules
  in QM7, obtained during five-fold cross-validation. The upper plot
  is a standard histogram. The lower plot displays an indication of
  the density function of the entire collection of contributions (the
  topmost object), as well as indications of the individual density
  functions of all contributions by atoms of a particular element
  (remaining objects). Straight lines indicate range, with mean
  marked; curves are kernel density estimators.}
\end{figure}

To inspect the distribution of the atomic contributions for
atomization energy, we repeated the computation for the
cross-validation of the complete QM7 dataset with the reduced Coulomb
descriptor. The full set of all atomic contributions for each
cross-validation split was retained, and the values for each of the
five splits were compiled; the resulting set can be considered as a
collection of all atomic contributions obtained during prediction of
that dataset. A histogram of the distribution of all atomic
contributions for QM7 can be found
in \Fref{fig:atomic_contribution_distribution}; additionally, a violin
plot displaying the distribution of all atomic contributions (plotted
by a kernel density estimator), as well as those of each of the five
atom types contained in the dataset (hydrogen, carbon, nitrogen,
oxygen, and sulfur) is given in the same figure.

The histogram displays a notable peak around approximately -70. This
peak can be understood by inspection of the violin plots for the
specific elements; a large majority of the contributions for hydrogen
atoms cluster around this value. This suggests that many of the
hydrogen atoms contribute similar amounts to the atomization energy of
their respective molecules, and is interesting, because hydrogen atoms
are by far the most prevalent in the QM7 dataset. By contrast, the
contributions for the other atom types are relatively spread out.

Although more work needs to be done with respect to the classification
of these contributions, we can draw one particular preliminary
conclusion from them. The tight clustering of hydrogen atoms suggests
large-scale redundancies of information in the underlying dataset; in
the context of the GAP framework, this will produce covariance
matrices with high condition numbers and corresponding numerical
inaccuracies. This is born out by our informal observations; we have
noted that covariance matrices encountered when attempting to train
potentials using even larger datasets than those used here (such as
the full GDB9 dataset) have prohibitively high condition, particularly
for suboptimal choices of kernel and descriptor parameters. We suspect
that methods for reducing the redundancies contained within the
training datasets (such as low-rank approximations of covariance
matrices and/or dataset sparsification techniques) will yield results
of significantly higher quality, while reducing computational cost.

\section{Conclusions and Future Work}
\label{sec:VMD_conclusions}

We have introduced three new numerical representations for atomic
neighborhoods, which can be applied to the Gaussian approximation
potential (GAP). Numerical experiments showed that our LC-GAP
implementation is capable of predicting the atomization energies of
organic molecules in the QM7 dataset at levels of accuracy competitive
with similar potentials already extant in the literature, for all
three presented representation types. Furthermore, we have
demonstrated that LC-GAP potentials trained on systems containing
lower numbers of atoms can be used to predict the energy values for
systems containing higher numbers of atoms, again with acceptable
accuracy.

Of the three representations presented here, the reduced Coulomb
descriptor has linear dimensionality in the number of particles in a
local neighborhood, while the other two descriptor types have
quadratic dimensionality. Despite this, the reduced Coulomb descriptor
performs comparably well, if not better, in all cases. This descriptor
also performs well when used to evaluate alternative molecular
properties than the atomization energy, producing results comparable
to the literature when applied to the QM7b many-properties
dataset. Finally, when used for cross-validation on the QM7b and GDB9
datasets, LC-GAP equipped with the reduced Coulomb descriptor predicts
atomization energies at and comfortably below chemical accuracy
respectively, achieving MAEs of approximately 1.00 and 0.78 kcal/mol.

A number of questions remain about the utility of these representation
functions when training on and predicting different kinds of atomic
systems. In particular, as all systems in the QM7 dataset are
equilibrium-state biomolecules, it would be interesting to investigate
the behavior of the LC-GAP system over non-equilibrium systems, as
well as infinitely-periodic crystal systems such as
semiconductors. Additionally, an analysis of per-atom contributions
produced during cross-validation suggest the existence of high levels
of redundancy in the QM7 dataset; we strongly suspect that the
application of techniques for lessening the impact of this redundancy
will produce worthwhile results.

\section*{Acknowledgement}
This work was funded in part by the German Federal Ministry for
Education and Research, under the Eurostars project E!6935
ATOMMODEL. We would also like to thank Maharavo Randrianarivony for
the fruitful discussion.

\bibliographystyle{abbrv}
\bibliography{author}

\begin{thebibliography}{10}

\bibitem{bartok2013}
A.~P. Bart{\'o}k, R.~Kondor, and G.~Cs{\'a}nyi.
\newblock On representing chemical environments.
\newblock {\em Physical Review B}, 87(18):184115, 2013.

\bibitem{bartok2010}
A.~P. Bart{\'o}k, M.~C. Payne, R.~Kondor, and G.~Cs{\'a}nyi.
\newblock Gaussian approximation potentials: The accuracy of quantum mechanics,
  without the electrons.
\newblock {\em Physical Review Letters}, 104(13):136403, 2010.

\bibitem{blum2009}
L.~C. Blum and J.-L. Reymond.
\newblock 970 million druglike small molecules for virtual screening in the
  chemical universe database {GDB-13}.
\newblock {\em Journal of the American Chemical Society}, 131(25):8732--8733,
  2009.

\bibitem{de2016comparing}
S.~De, A.~P. Bart{\'o}k, G.~Cs{\'a}nyi, and M.~Ceriotti.
\newblock Comparing molecules and solids across structural and alchemical
  space.
\newblock {\em Physical Chemistry Chemical Physics}, 18(20):13754--13769, 2016.

\bibitem{filszar1994}
S.~Flisz{\'a}r.
\newblock {\em Atoms, Chemical Bonds, and Bond Dissociation Energies}.
\newblock Lecture Notes in Chemistry. Springer-Verlag, 1994.

\bibitem{griebel2007}
M.~Griebel, S.~Knapek, and G.~Zumbusch.
\newblock {\em Numerical Simulation in Molecular Dynamics: Numerics,
  Algorithms, Parallelization, Applications}, volume~5 of {\em Texts in
  Computational Science and Engineering}.
\newblock Springer Science \& Business Media, Heidelberg, 2007.

\bibitem{hansen2015}
K.~Hansen, F.~Biegler, R.~Ramakrishnan, W.~Pronobis, O.~A. von Lilienfeld,
  K.-R. Müller, and A.~Tkatchenko.
\newblock Machine learning predictions of molecular properties: Accurate
  many-body potentials and nonlocality in chemical space.
\newblock {\em The Journal of Physical Chemistry Letters}, 6(12):2326--2331,
  2015.

\bibitem{hansen2013}
K.~Hansen, G.~Montavon, F.~Biegler, S.~Fazli, M.~Rupp, M.~Scheffler, O.~A. von
  Lilienfeld, A.~Tkatchenko, and K.-R. M\"uller.
\newblock Assessment and validation of machine learning methods for predicting
  molecular atomization energies.
\newblock {\em Journal of Chemical Theory and Computation}, 9(8):3404--3419,
  2013.

\bibitem{kohn1996}
W.~Kohn.
\newblock Density functional and density matrix method scaling linearly with
  the number of atoms.
\newblock {\em Physical Review Letters}, 76(17):3168, 1996.

\bibitem{montavon2012}
G.~Montavon, K.~Hansen, S.~Fazli, M.~Rupp, F.~Biegler, A.~Ziehe, A.~Tkatchenko,
  O.~A. von Lilienfeld, and K.-R. M{\"u}ller.
\newblock Learning invariant representations of molecules for atomization
  energy prediction.
\newblock In {\em Advances in Neural Information Processing Systems}, pages
  440--448, 2012.

\bibitem{montavon2013}
G.~Montavon, M.~Rupp, V.~Gobre, A.~Vazquez-Mayagoitia, K.~Hansen,
  A.~Tkatchenko, K.-R. M{\"u}ller, and O.~A. von Lilienfeld.
\newblock Machine learning of molecular electronic properties in chemical
  compound space.
\newblock {\em New Journal of Physics}, 15(9):095003, 2013.

\bibitem{plimpton2012}
S.~J. Plimpton and A.~P. Thompson.
\newblock Computational aspects of many-body potentials.
\newblock {\em MRS bulletin}, 37(05):513--521, 2012.

\bibitem{prodan2005}
E.~Prodan and W.~Kohn.
\newblock Nearsightedness of electronic matter.
\newblock {\em Proceedings of the National Academy of Sciences of the United
  States of America}, 102(33):11635--11638, 2005.

\bibitem{ramakrishnan2014}
R.~Ramakrishnan, P.~O. Dral, M.~Rupp, and O.~A. von Lilienfeld.
\newblock Quantum chemistry structures and properties of 134 kilo molecules.
\newblock {\em Scientific Data}, 1, 2014.

\bibitem{rasmussen2006}
C.~E. Rasmussen and C.~K.~I. Williams.
\newblock {\em Gaussian Processes for Machine Learning}.
\newblock The MIT Press, 2006.

\bibitem{rupp2015}
M.~Rupp.
\newblock Machine learning for quantum mechanics in a nutshell.
\newblock {\em International Journal of Quantum Chemistry}, 115(16):1058--1073,
  2015.

\bibitem{rupp2015b}
M.~Rupp, R.~Ramakrishnan, and O.~A. von Lilienfeld.
\newblock Machine learning for quantum mechanical properties of atoms in
  molecules.
\newblock {\em The Journal of Physical Chemistry Letters}, 6(16):3309--3313,
  2015.

\bibitem{rupp2012}
M.~Rupp, A.~Tkatchenko, K.-R. M{\"u}ller, and O.~A. von Lilienfeld.
\newblock Fast and accurate modeling of molecular atomization energies with
  machine learning.
\newblock {\em Physical Review Letters}, 108(5):058301, 2012.

\bibitem{tadmor2011}
E.~Tadmor and R.~Miller.
\newblock {\em Modeling Materials: Continuum, Atomistic and Multiscale
  Techniques}.
\newblock Cambridge University Press, 2011.

\bibitem{willmott2005}
C.~J. Willmott and K.~Matsuura.
\newblock Advantages of the mean absolute error ({MAE}) over the root mean
  square error ({RMSE}) in assessing average model performance.
\newblock {\em Climate Research}, 30(1):79, 2005.

\end{thebibliography}

\end{document}